\documentstyle[aclap,fancyheadings,psfig]{article}

\title{A Divide-and-Conquer Strategy for Parsing}
\author{Peh Li Shiuan \\
Defence Science Organization \\
20 Science Park Drive \\
Singapore 118230 \\
Republic of Singapore \\ 
{\em plishiua@trantor.dso.gov.sg}
\And
Christopher Ting Hian Ann \\
Computational Science Programme \\
National University of Singapore \\
10 Kent Ridge Crescent \\
Singapore 119260 \\
Republic of Singapore \\
{\em cting@galaxy.cz3.nus.sg}}
\begin{document}
\thispagestyle{fancy}
\headrulewidth 0 pt
\chead{\em In Proceedings of the ACL/SIGPARSE 5th International Workshop on Parsing Technologies,\\ Santa Cruz, USA, pp. 57 -- 66.}
\maketitle

\begin{abstract}
In this paper, we propose a novel strategy which is designed to enhance the
accuracy of the parser by simplifying complex sentences before parsing.
This approach involves the separate parsing of the constituent 
sub-sentences within a complex sentence.
To achieve that, the divide-and-conquer strategy first disambiguates the roles
of the {\em link words} in the sentence and segments the sentence based on
these roles. The separate parse trees of the segmented sub-sentences and
the noun phrases within them are then synthesized to form the final parse.
To evaluate the effects of this strategy on parsing, we compare the original
performance of a dependency parser with the performance when it is
enhanced with the divide-and-conquer strategy. 
When tested on 600 sentences of the IPSM'95 data sets,
the enhanced parser saw a considerable error reduction
of 21.2\% in its accuracy.
\end{abstract}

\bibliographystyle{acl}
\section{Introduction}

In Black's survey of the state of the art in parsing, she picked several 
features which she considered as relevant in categorizing the type of data
used in the evaluation of parsers' performance \cite{Black:state-of-the-art}.
One of these, notably, is the length of the test sentences. 
The reason behind sentence length being a discriminating factor 
is the typical performance degradation of parsers as test 
sentences increase in length and complexity.

Most parsers circumvent this performance degradation by catering 
for complex sentence contructs specifically within the parsing mechanism. 
However, whilst the complexity of the parser is greatly increased, the
resulting parsing accuracy still leaves much to be desired.

In this paper, we propose a novel approach to enhance the parsing of long,
complex sentences. Instead of devising ways to improve the parser itself,
our divide-and-conquer strategy attempts to enhance parsing through the 
simplification of the inputs to the parser.

In brief, the divide-and-conquer strategy involves the separate parsing of 
the constituent sub-sentences within a complex sentence. To achieve that,
the divide-and-conquer strategy first examines the link words\footnote{Link 
words refer to punctuations, conjunctions and prepositions.} of a sentence and
disambiguates their specific roles in the sentence. Based on these roles, it is
determined if the link words form segmentation points of the sentence. The
segmented sub-sentences and the extracted noun phrases are then parsed 
separately. Finally, the separate parse trees are fused together to form a
complete parse. Figure~1 illustrates the flow-of-processing
of the divide-and-conquer strategy with a test sentence.

\begin{figure}[tp]
\centerline{\hbox{\psfig{figure=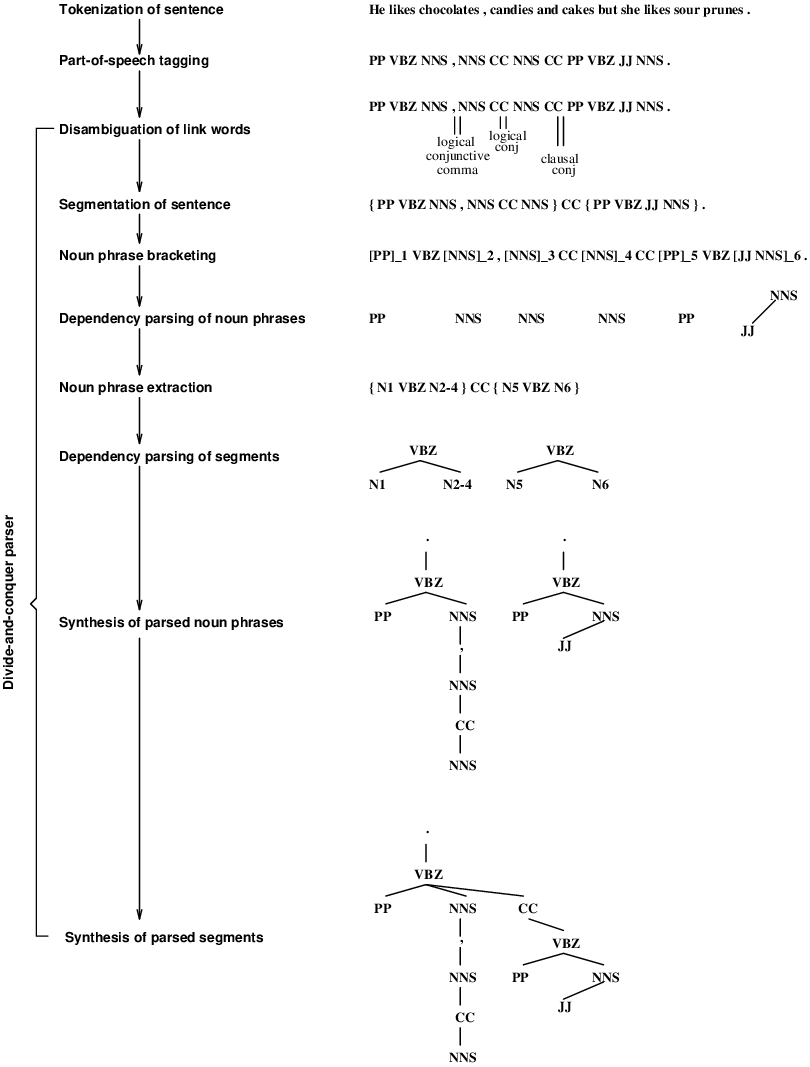}}}
\hfill
\begin{center}
\parbox{\textwidth}{
{\bf Figure~1. 
An overview of the flow of processing of the divide-and-conquer parser.}
After tokenization and part-of-speech tagging, the link words 
in the sentence are disambiguated. During the segmentation phase, the sentence
is segmented at the clausal conjunction {\sf ``but"}. 
Thereafter,
noun phrase bracketing is done, and all the noun phrases ({\sf ``He"},
{\sf ``chocolates"}, {\sf ``candies"}, {\sf ``cakes"}, {\sf ``she"}, 
{\sf ``sour prunes"}) 
are parsed separately. To simplify the segments further, noun phrases and 
noun phrase groups are extracted from the segments
and replaced with a single noun.
Finally, the separate parse trees of
the noun phrases are glued onto the segments' parse trees and the segments'
parse trees are synthesized to form the final parse tree of the sentence.
}
\end{center}
\end{figure}

The parser used in this work is a dependency structure parser. However, the
divide-and-conquer strategy can similarly be adapted to a constituency 
parser\footnote{Only the synthesis engine needs to be modified to cater for 
constituency structures.}.
This is because both syntactic structures exhibit a property which endear them
to the divide-and-conquer approach.
For both, the parse structure of a sentence is actually made up of 
sub-structures which are themselves legal parse structures of the constituent
sub-sentences. Figures~2 and~3 
highlight this property of dependency and constituency structures respectively.
Mel'\v{c}uk discusses these two prongs of syntactic representation in more detail
\cite{Melcuk:dep}.

\begin{figure}[tp]
\centerline{\hbox{\psfig{figure=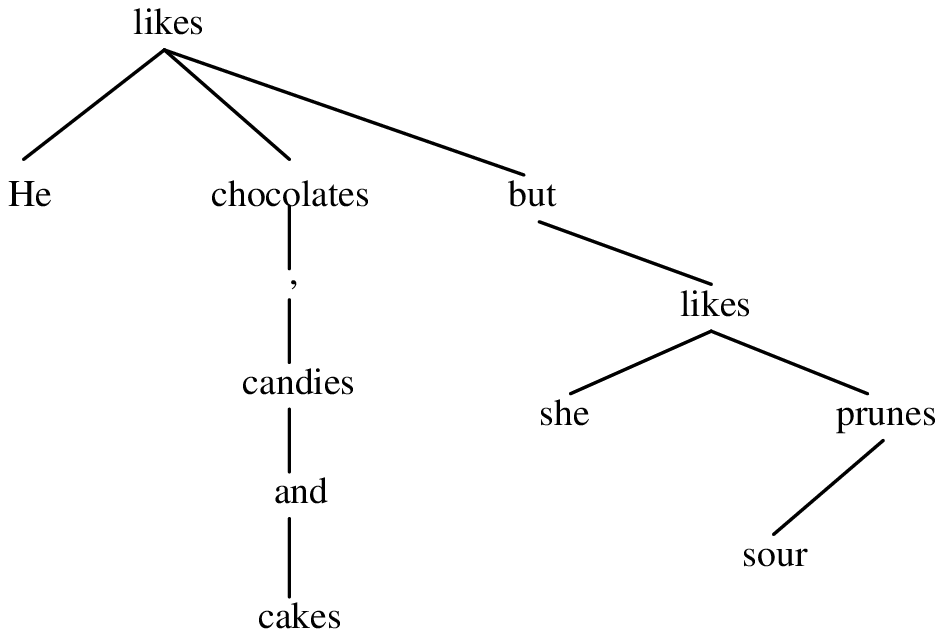}}}
\hfill
\begin{center}
\parbox{\textwidth}{
{\bf Figure~2.
The dependency structure of {\sf ``He likes chocolates, candies and cakes but
she likes sour prunes."}.}
As is evident, the dependency structure is actually composed of two sub-trees
linked via the conjunction {\sf ``but"}.
One sub-tree is the dependency structure for {\sf ``He likes chocolates, 
candies and cakes."} whilst the other is the parse tree of 
{\sf ``She likes sour prunes."}.
}
\end{center}
\end{figure}

\begin{figure}[tp]
\centerline{\hbox{\psfig{figure=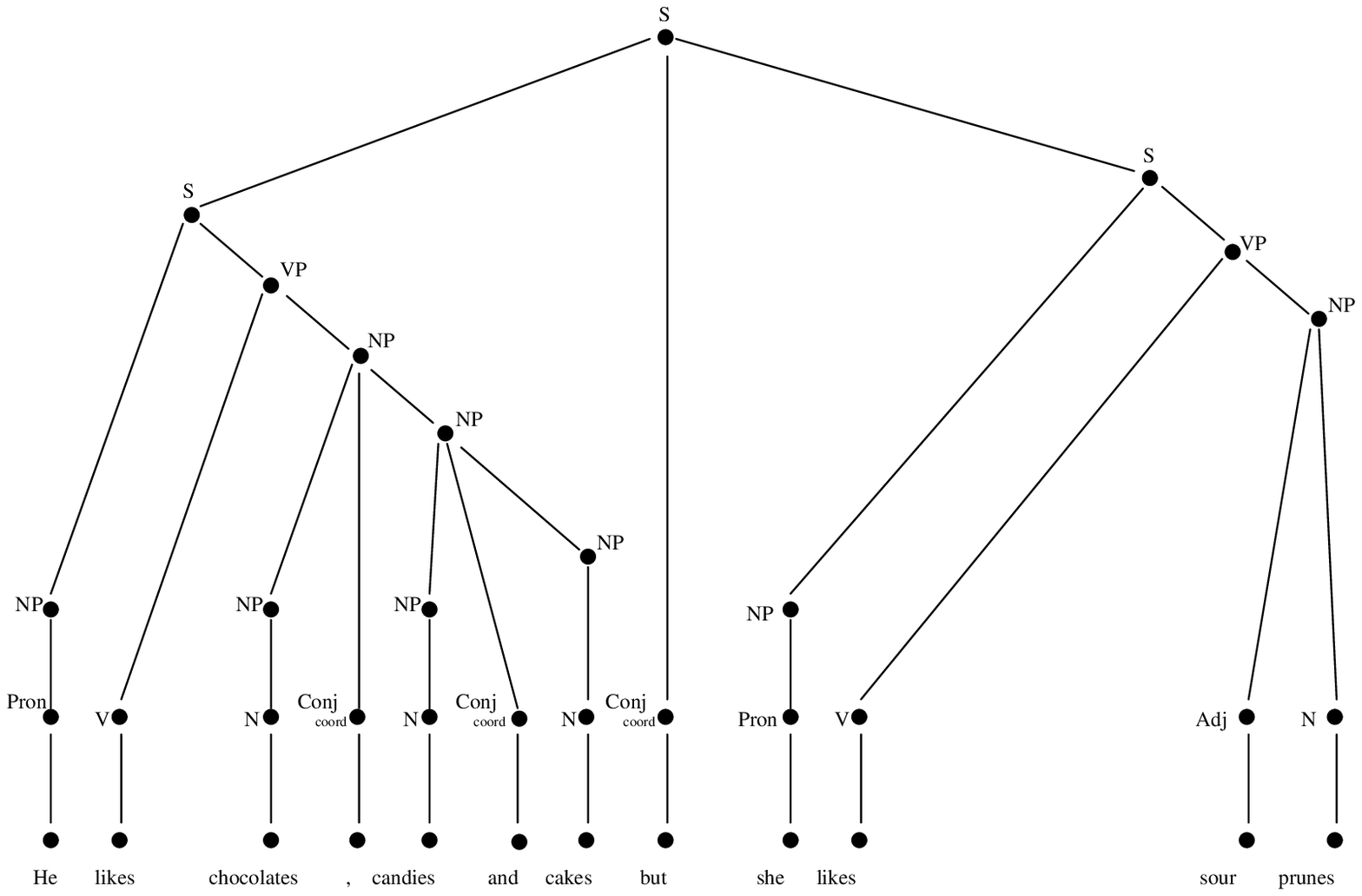}}}
\hfill
\begin{center}
\parbox{\textwidth}{
{\bf Figure~3.
The constituency structure of {\sf ``He likes chocolates, candies and cakes but
she likes sour prunes."}.}
Here, the constituency parse tree is clearly made up of the two parse trees
of the sub-sentences and the coordinating conjunction {\sf ``but"}. 
(The structure notations used here adhere to those in \cite{Melcuk:dep}.)
}
\end{center}
\end{figure}

To investigate the effects of this strategy on parsing, we compare the
original performance of our dependency 
parser \cite{Ting+How:ipsm} with the
performance when it is enhanced with the divide-and-conquer strategy.
As we are unaware of comparable approaches in literature, we are unable to
provide comparisons of the resultant improvement in performance. However,
the 21.2\% error reduction in parsing accuracy, in our opinion, is encouraging.

Following, Section~\ref{sec-motivation} illustrates the shortcoming faced by 
the original parser.
Section~\ref{sec-description} then describes each phase of the 
divide-and-conquer strategy in detail and past related efforts are chronicled
in Section~\ref{sec-related}. Section~\ref{sec-performance} presents the
performance evaluation results and Section~\ref{sec-conclusion} concludes
the paper.

\section{Motivation}
\label{sec-motivation}

The original dependency parser, our baseline for comparison in this paper, 
uses an enhanced Hidden Markov Model \cite{Ting+How:ipsm}.
In this statistical approach, each word can depend on every other word in the 
sentence.
The task of the enhanced Hidden Markov Model is to choose the
most likely governor for each word.
As the length of a sentence increases, the number of possible governors of
each word increases too. Parsing accuracy thus deteriorates.

This shortcoming of the parser was the prime motivation behind the
divide-and-conquer strategy. By reducing the length of the input sentence, 
the statistical perplexity of the parsing problem greatly reduces.
One would expect improvement in the parsing accuracy. 

For instance, if the sentence {\sf ``He likes oranges but she prefers apples.''}
is parsed by the original parser, the word {\sf ``oranges''} can potentially
be a dependent of any other word in the sentence -- {\sf ``He'', ``likes'',
``she'', ``prefers'', ``apples''}. However, with the divide-and-conquer
strategy, once the sentence is segmented into {\sf ``He likes oranges.''} and
{\sf ``She prefers apples.''}, the word {\sf ``oranges''} 
has only two potential governors, {\sf ``He''} and 
{\sf ``likes''}. It is hence evident that the statistical perplexity is 
lower with shorter sentences.

\section{Description}
\label{sec-description}

\subsection{Disambiguation of link words}
\label{subsec-disamb}

The disambiguation phase is responsible for identifying the specific roles of
link words in a sentence. These roles offer strong clues as to the boundaries
of the sub-sentences.

However, due to the wide variety of ways in which link words are used in the 
English language, linguistic experts have, in the past, been able to identify 
ten to twenty uses of commas alone \cite{Nunberg:punc}. 
To enhance the effectiveness of the disambiguation, 
we sieved out just two roles of commas - prosodic and conjunctive
and two roles of conjunctions - logical and clausal.
(Prepositions are disambiguated using only the
wordform of the preposition. For instance, prepositions such as 
``although'', ``if'' and ``while'' are classified as subordinating 
prepositions.)

Simply put, a prosodic comma is one which marks
a pause in the sentence whilst a conjunctive comma links parts
of a sentence together. An example will clarify their differences. In 
{\sf ``When Jane goes to school, she takes a bus, walks 5 minutes and continues
the journey on the rail."}, 
the first comma is a prosodic one since it indicates a prosodic
break. The second comma, however, is conjunctive and can be easily substituted 
with a conjunction. Naturally, a conjunctive comma takes on the roles of
conjunctions too.

A logical conjunction is one which links\footnote{The link in this context 
refers to the dependency links between words in a sentence. Hence, if a
conjunction links nouns, it means that it depends on a noun and another noun
depends on the conjunction itself.} similar components or items such as 
nouns, adjectives, adverbs and so on. 
A clausal conjunction, on the other 
hand, joins sub-sentences together. The examples given below
will reveal the subtle difference between these two roles. 
An example of a logical
conjunction is in {\sf ``I like ice-cream, hot-dogs but not pies."}. The 
same conjunction {\sf ``but"}, however, is clausal in {\sf ``I like 
ice-cream, crave for hot-dogs but detest pies."}. The distinct difference
between these two roles is that a clausal conjunction 
links verbs instead of nouns, adjectives, adverbs {\it etc.}. 
 
The disambiguation mechanism employed is a neural network\footnote{We 
explored the nearest neighbour algorithm as an alternative
disambiguation mechanism. Both algorithms achieve similar accuracy performance.
Neural networks were chosen mainly because testing time is on the critical
path of the parsing process.
}.
Figure~4 shows the neural network model used. 
The inputs to the network are the neighbouring parts of speech of the sentence, 
since the roles of commas and conjunctions depend greatly on the sentence 
context they
are in. The output of the network is simply a classified role of the comma or
conjunction.

\begin{figure}[htp] \begin{center}
\centerline{\hbox{\psfig{figure=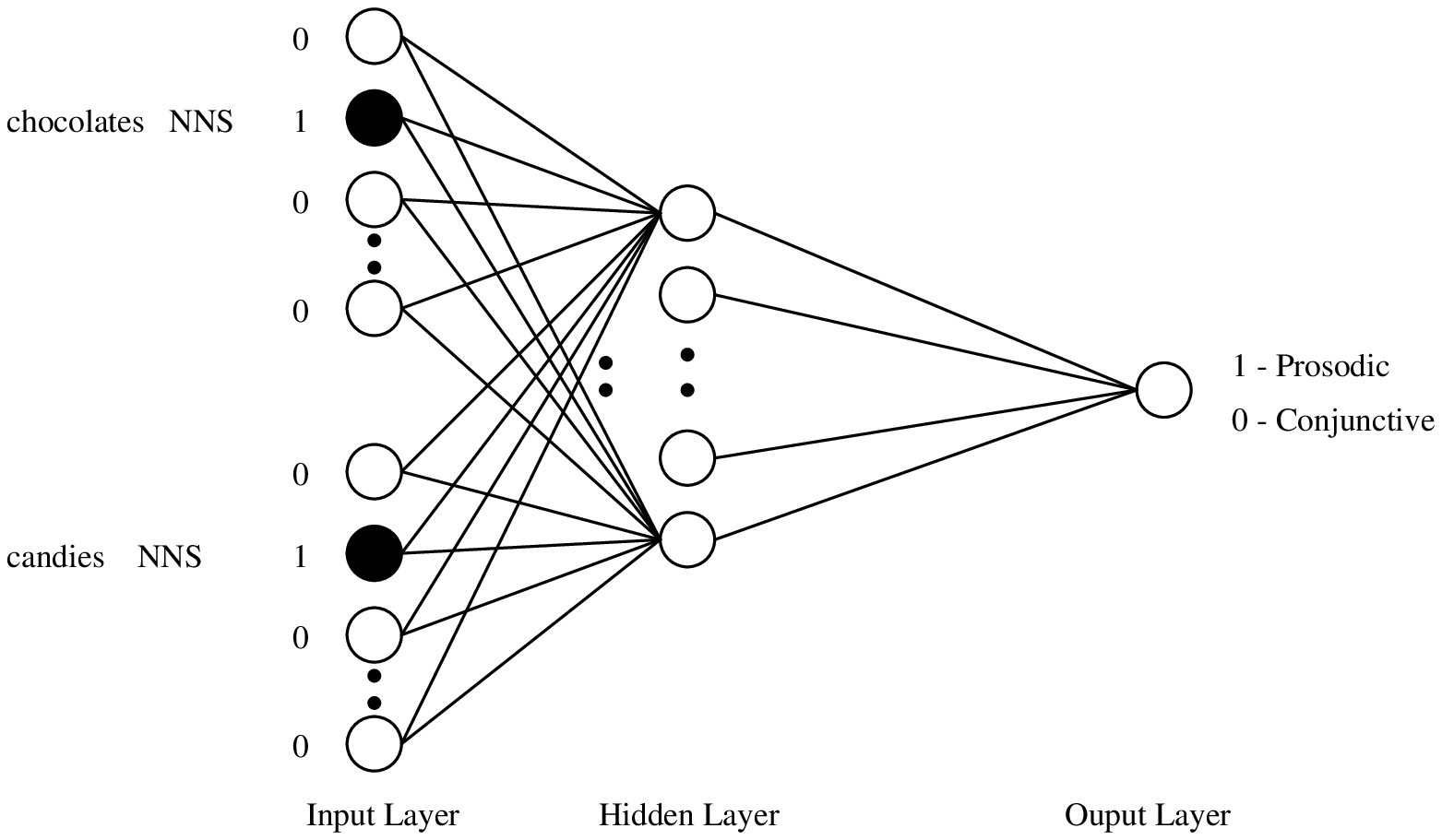,scale=75}}}
\hfill
\end{center}
\begin{center}
\parbox{\textwidth}{
{\bf Figure~4. Modelling of disambiguation process using neural
network.}
The same test sentence as Figure~4 - {\sf ``He likes 
chocolates, candies and cakes ..."} is used. If the window size is two, i.e
. we only use the two words before and after the comma as our inputs in comma 
disambiguation ({\sf ``chocolates"} and {\sf ``candies"}), the neural network
used will be as above. It has two sets of input nodes for the two words, and 
only the input node which corresponds to the part of speech of the word is set
to 1. The rest of the input nodes are set to 0. The network will then
determine the classification of the comma and output 1 or 0 accordingly.
}
\end{center}
\end{figure}

\subsection{Segmentation of sentence}

The segmentation phase uses the properties of the roles of link words to
pin-point the exact segmentation points in the sentence. A sentence is
segmented at the link word if the link word is a prosodic or clausal
conjunctive comma, a clausal conjunction or a subordinating preposition.

\subsection{Noun phrase backeting}

We developed a noun phrase parser \cite{Ting:npparser} for our noun phrase
bracketing module.
It is based on an enhanced HMM model and is reported to achieve 96.3\% accuracy 
(exact match of pairs of bracketings) on the WSJ Corpus.

\subsection{Dependency parsing of noun phrases}

In the parsing of noun phrases, the original statistical parser, which is also 
based
on an enhanced HMM model, is trained on noun phrases instead of whole sentences.
This is a simplification of the statistical parser, 
analogous to retaining only those rules in a rule-based parser which are 
relevant to noun phrases. 

\subsection{Noun phrase extraction}

Each segment can be further compressed by replacing its noun phrases and
noun phrase groups with a single node. 
Noun phrase groups refer to a string of noun phrases
delimited by logical conjunctive commas, logical conjunctions or 
prepositions such as {\sf ``of"}. Hence, after the noun phrase extraction
phase, the sentences {\sf ``He likes chocolates, candies and cakes."}, 
{\sf ``The cat likes fish."} and {\sf ``The President of the United States of
America meets the Queen of England."} will all yield {\sf ``NN VBZ NN."}
(see \cite{Marcus+al:penntree} for an explanation of the 
notation symbols of parts of speech used in this paper.) 

\subsection{Dependency parsing of segments}

For parsing of segments, we simplified the parser by training it on 
sub-sentences instead of whole sentences. This is again analogous to 
removing those rules in a rule-based parser which cater for noun phrase
groups and complex constructs, thus reducing the complexity of the parser
and the number of possible parses generated.

\subsection{Synthesis of parsed noun phrases}

To attach the separate parse trees of each noun phrase back onto
the parse tree of the segment, each noun phrase is made dependent on the 
governor of the single node which represented it.
For noun phrase groups, the noun phrases
in the group are chained one after another, with the first noun phrase
being the head of the entire noun phrase group.

\subsection{Synthesis of parsed segments}

The welding of the parse trees of segments is performed using a
rule-based synthesis engine\footnote{Rules were
designed based on the Brown Corpus.}. 
This synthesis engine first connects the link 
words, {\it i.e.} the prosodic commas, the clausal conjunctions and the 
subordinating prepositions, to the correct words in the segment parse trees. 
It then establishes the dependencies between the segments and completes
the final parse tree.

The synthesis of prosodic commas is trivial due to the convention adopted
by the original parser, whereby
all prosodic commas are leaf nodes which depend on the word just before it. 

{\bf Rules :-}

Let a sentence be represented as \{ {\em ($segment_{0}$) $linkword_{0}$ ($segment_{1}$) $linkword_{1}$ ... ($segment_{i}$) $linkword_{i}$ ($segment_{i+1}$) $linkword_{i+1}$ ... ($segment_{n}$) $linkword_{n}$} \}

{\bf If} $linkword_{i}$ is a prosodic comma {\bf then}

\begin{tabbing}
Blankspace: \= blah blah \kill
            \> $linkword_{i}.governor$ = last word in $segment_{i}$ \\
            \> $linkword_{i}.dependent$ = NULL
\end{tabbing}

As mentioned in Section~\ref{subsec-disamb}, clausal conjunctions link
verbs. Hence, during the synthesis of clausal conjunctions, the task
of the synthesis engine is to find their respective governor and dependent 
verbs.
The key feature of clausal conjunctions is that they
link segments which are similar.
Our synthesis engine thus exploits three syntactic clues to determine the
similarity between potential verbs --
the tense of the verb, the morphological form of the verb and the word 
position of the verb.

{\bf Rules :-}

{\bf If} $linkword_{i}$ is a clausal conjunctive comma {\bf or} a clausal conjunction {\bf then}

\begin{tabbing}
Blankspace: \= {\bf Else} \= {\bf For} \= {\bf Else} \= \kill
	    \> {\bf If}   \> number of words in $segment_{i}$ is zero {\bf then} \> \> \\
	    \>            \> $linkword_{i}.governor$ = leftmost verb in $segment_{i+1}$ \> \> \\
	    \>            \> $linkword_{i}.dependent$ = NULL \> \> \\
	    \> {\bf Else} \> \> \> \\
	    \>            \> $linkword_{i}.dependent$ = head verb of nearest segment to the right of $segment_{i}$ \> \> \\
	    \>            \> {\bf For} \> $j$ {\bf from} $i-1$ {\bf to} $0$ \> \\
	    \>            \>           \> {\bf If} \> morphological form of head verb of $segment_{j}$ = \\
	    \>            \>           \>          \> morphological form of $linkword_{i}.dependent$ {\bf or} \\
	    \>            \>           \>          \> tense of head verb of $segment_{j}$ = \\
            \>            \>           \>          \> tense of $linkword_{i}.dependent$ {\bf then} \\ 
	    \>            \>           \>          \> $linkword_{i}.governor$ = head verb of $segment_{j}$ \\
	    \>	   	  \>           \> {\bf Else} \> \\
	    \> 		  \>           \>            \> $linkword_{i}.governor$ = rightmost verb in $segment_{i}$ 
\end{tabbing}

As for subordinating prepositions, more specific rules which cater for the
different prepositions are derived. For instance, the synthesis of the 
preposition {\sf ``that"} is rather similar to that of clausal  
conjunctions, as in this sentence {\sf ``I know that he is angry."}, where
{\sf ``that"} depends on the verb {\sf ``know"} and {\sf ``is"} depends on
{\sf ``that"}. However, {\sf ``that"} can also depend on adjectives, such as
in {\sf ``I am glad that I have gained weight."}, where {\sf ``that"} depends 
on the adjective {\sf ``glad"}.
After all these link words have been connected by the synthesis engine, 
it has to tie up the remaining loose ends. Firstly, the engine needs to
identify the head segment, {\it i.e.}, the segment which depends on no other segments.

{\bf Rules :-}

\begin{tabbing}
Blankspace: \= {\bf For} \= {\bf If} \= blah \kill
            \> {\bf For} $j$ {\bf from} $0$ {\bf to} $n$ \> \> \\
	    \>           \> {\bf If} head verb of $segment_{j}$ do not have a governor {\bf and} \> \\
	    \>           \>       \> head verb of $segment_{j}$ is not a continuous verb {\bf and} \\
	    \>           \>       \> first word of $segment_{j}$ is not ``when", ``while", ``also", ``until", ``to", {\bf then} \\
	    \>           \>          \> governor of head verb of $segment_{j}$ = $linkword_{n}$ 
\end{tabbing}

Then, for segments which are not yet linked, they are chained to the head
segment in order.

\section{Related Work}
\label{sec-related}

Several works in literature specifically target complex sentences linked via
link words in their attempt to improve parsing. All differ from the 
divide-and-conquer strategy in that they involve enhancements to the parsing
mechanism itself.

Magerman discussed the poor performance of his parser SPATTER on sentences
with conjunctions \cite{Magerman:spatter}.
As a result, he augmented SPATTER's probabilistic model
with an additional conjunction feature. However, he reported that though
SPATTER's performance on conjoined sentences improves with the conjunction
feature, a significant percentage is still misanalyzed, as the simple
conjunction feature model finds it difficult to capture long distance 
dependencies.

Kurohashi and Nagao developed a method which was geared towards conjoined
sentences too \cite{Kurohashi+Nagao:conj}. 
Their approach involves the use of a language-dependent 
conjunction scoping method to detect conjunctive structures in Japanese 
sentences. They then modify a rule-based
dependency parser to cater for this additional information on conjunctive
structures. 

Jones explored another type of link words, the punctuations 
\cite{Jones:punc}.
He showed successfully that for longer sentences, 
a grammar which makes use of punctuation massively outperforms one which does 
not.
Besides improving parsing accuracy, the use of punctuations also 
significantly reduces the number of possible parses generated.
However, as theoretical forays into the syntactic roles of punctuation are
limited, the grammar he designed can only cover a subset of all punctuation
phenomena. Unexpected constructs thus cause the grammar to fail completely.

The importance of punctuation in reducing syntactic ambiguity is further
attested to by Briscoe and Carroll \shortcite{Briscoe+Carroll:punc}. They 
conducted parsing experiments on identical texts with and without 
punctuation marks and showed the clear improvement in parsing performance.
However, they highlighted commas as a major source of ambiguity in their 
analysis and indicated that syntactic context may be necessary for the effective
disambiguation of comma.

All these approaches illustrate that parsing accuracy do improve with 
additional information on link words. However, all were faced with the 
non-trivial task of incorporating this information directly into their
respective parsing mechanisms.

\section{Performance Evaluation}
\label{sec-performance}

We conducted experiments on 3 sets of data used in the IPSM'95 Workshop 
\cite{Ting+Peh:ipsm}, Dynix, Lotus and Trados. These data sets were
collected from software manuals and each comprises 200 sentences. The
length of the sentences in the data sets ranges from 2 to 59 words and
the average length is 17.4 words.
The tagger is trained on the PennTree Bank's Brown Corpus, the Wall Street 
Journal Corpus and the IPSM'95 Corpus.
The training arrangement for both the original parser and the 
divide-and-conquer parser is as follows. 
When the test data set is, say, Dynix,
the parser is trained on the other two data sets, Lotus and Trados, and
an additional 1421 sentences obtained from other software manuals and a subset
of the Brown Corpus. The noun phrase parser is trained with the same
arrangement as the parser.
The disambiguation mechanism, the neural network, is trained
on a small subset of the Brown Corpus. The window size used is 8 words.

The performance results of all 3 data sets are summarized in Table~1. 
Accuracy figures for 
part-of-speech tagging and parsing refer to word-level performance
, i.e. the number of words which are assigned with the correct part of speech
and the number of words which are linked to the right governor 
respectively\footnote{ We manually annotated the parse trees of the entire 
data set and compared 
the results with this corpus. Hence, even though certain sentences may have a
number of acceptable parses, only the one which matches the corpus exactly
is considered correct.}. The performance of the noun phrase parser refers to
the exact match of pairs of bracketings.

\hfil
{\small
\begin{center}
\begin{tabular}[hbt]{|l|c|c|c|c|} \hline
Divide-\&-Conquer Components & {\sf Dynix} & {\sf Lotus} & {\sf Trados} & Average \\ \hline
Part-of-Speech Tagging     & 96.3 \% & 94.8 \% & 97.3 \% & 96.1 \% \\ \hline
Comma Disambiguation       & 97.2 \% & 96.4 \% & 86.3 \% & 93.3 \% \\ \hline 
Conjunction Disambiguation & 96.8 \% & 92.7 \% & 91.7 \% & 93.7 \% \\ \hline
Noun Phrase Parsing        & 98.0 \% & 94.0 \% & 99.0 \% & 97.0 \% \\ \hline
\end{tabular}
\end{center}
}

\hfill
\hfill
\hfill
{\small
\begin{center}
\begin{tabular}[hbt]{|c|c|c|c|c|c|c|c|} \hline
\multicolumn{4}{|c|}{Original Parser} &
\multicolumn{4}{c|}{Divide-\&-Conquer Parser} \\ \hline
{\sf Dynix} & {\sf Lotus} & {\sf Trados} & Average &
{\sf Dynix} & {\sf Lotus} & {\sf Trados} & Average \\ \hline
82.5 \% & 80.9 \% & 79.9 \% & 81.1 \% & 
86.4 \% & 85.4 \% & 83.6 \% & 85.1 \% \\ \hline
\end{tabular}
\end{center}
}

\hfill
\begin{center}
\hfil
\centerline{\bf Table~1. Performance results}
\end{center}

The divide-and-conquer strategy turns in encouraging results. It improves
average word level parsing accuracy from 81.1\% to 85.1\% 
(an error reduction of 21.2\%).
It is also worth mentioning that this is achieved despite the small training 
corpus (1812 sentences) involved.

Figures~5 and~6 contrasts the resultant 
parse trees of the original and enhanced parsers on a test sentence. 
Figure~7 shows an example where errors in the disambiguation
process propagated and resulted in errors in the final parse tree.
The attachment of link words, as shown in Figure~7, is
affected by the performance 
of the disambiguation phase, which in turn is affected by the tagger. 

\begin{figure}[tp] \begin{center}
\centerline{\hbox{\psfig{figure=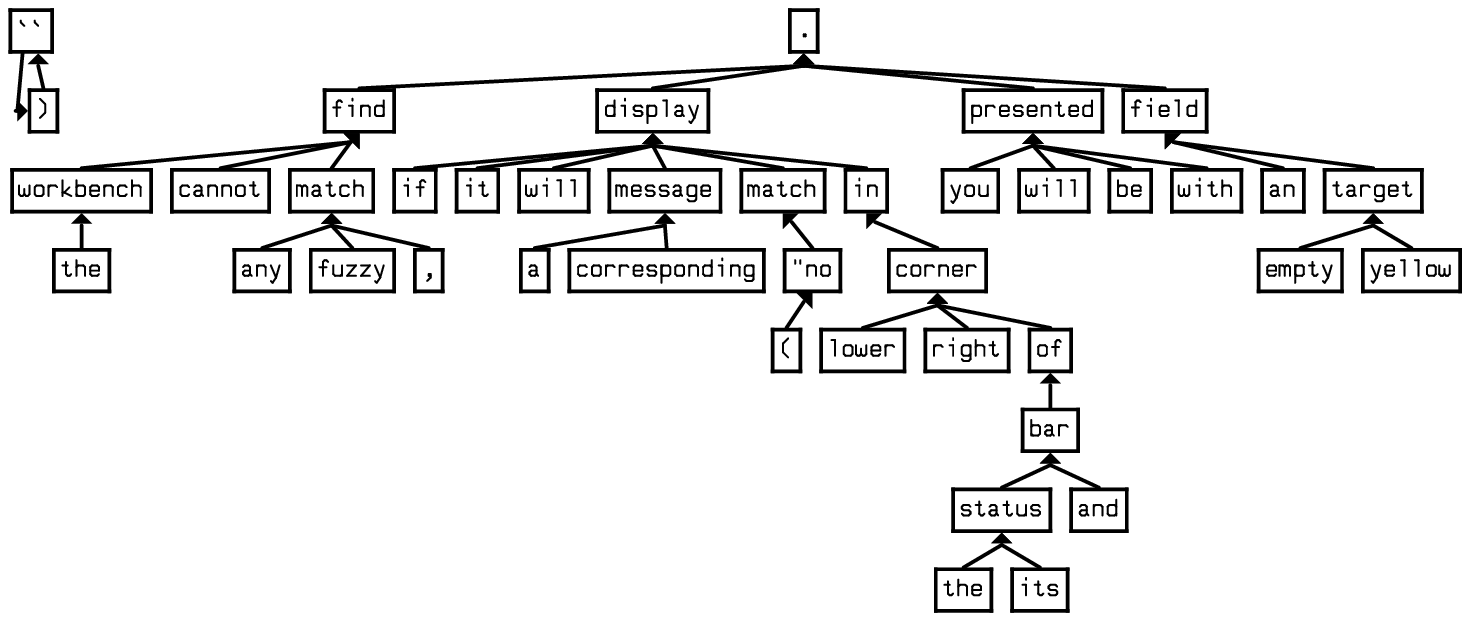}}}
\hfill
\end{center}
\begin{center}
\parbox{\textwidth}{
{\bf Figure~5. The parse tree produced by the original parser
for this sentence in the Trados data set -- {\sf ``If the Workbench cannot
find any fuzzy match, it will display a corresponding message ( ``No match ") in
the lower right corner of its status bar and you will be presented with an 
empty yellow target field."}.}
Here, the link words {\sf ``if"} and {\sf ``and"} are attached wrongly. Also,
the parser is unable to determine the top head verb to be {\sf ``display"}.
}
\end{center}
\end{figure}

\begin{figure}[tp] \begin{center}
\centerline{\hbox{\psfig{figure=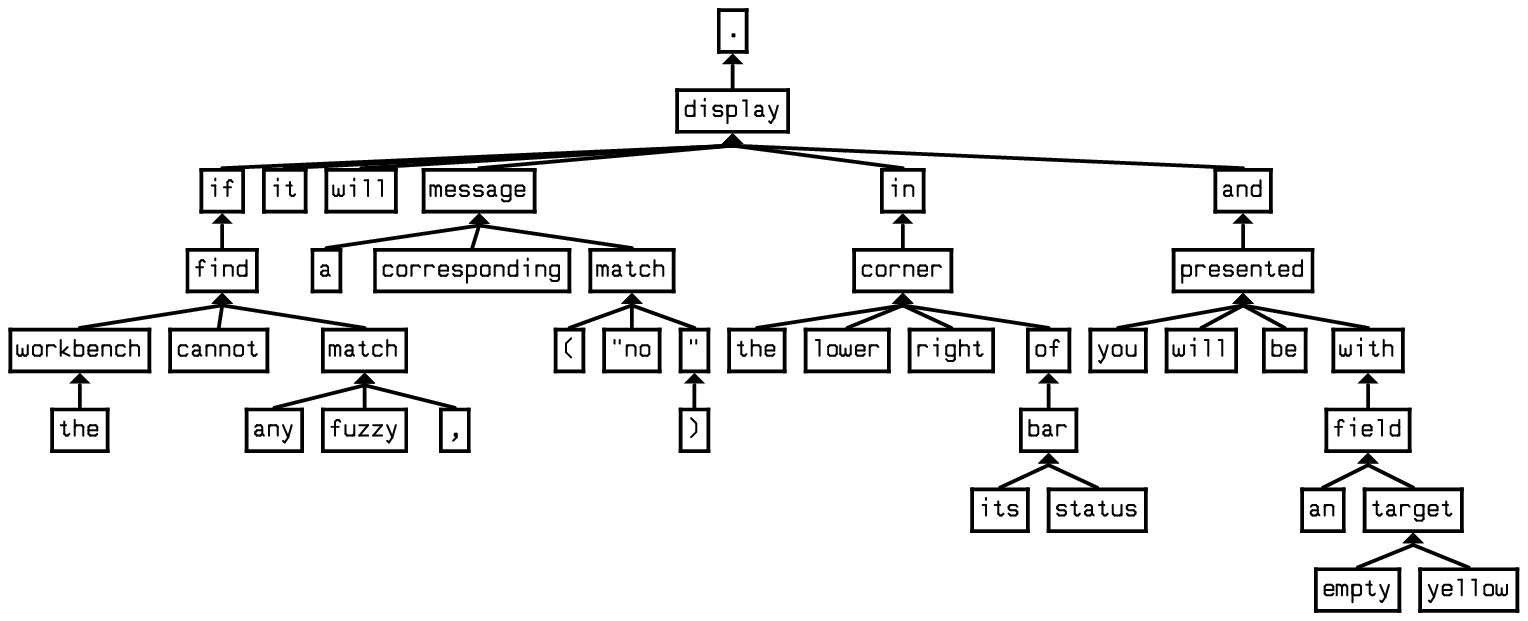}}}
\hfill
\end{center}
\begin{center}
\parbox{\textwidth}{
{\bf Figure~6. The parse tree produced by the parser enhanced
with the divide-and-conquer strategy on the same sentence in 
Figure~5.}
Here, the disambiguation phase manages to disambiguate all the link words
accurately. {\sf ``if"} is correctly identified as a subordinating preposition, 
the comma as a prosodic comma and {\sf ``and"} as a clausal conjunction. As
a result, the segmentation phase correctly segments the sentence into
{\sf ``The Workbench cannot find any fuzzy match."}, {\sf ``It will display a
corresponding message (``No match") in the lower right corner of its status
bar."} and {\sf ``You will be presented with an empty yellow target field."}.
The synthesis engine then attaches the link words correctly and is able to
identify {\sf ``display"} as the head verb.
}
\end{center}
\end{figure}

\begin{figure}[tp] \begin{center}
\centerline{\hbox{\psfig{figure=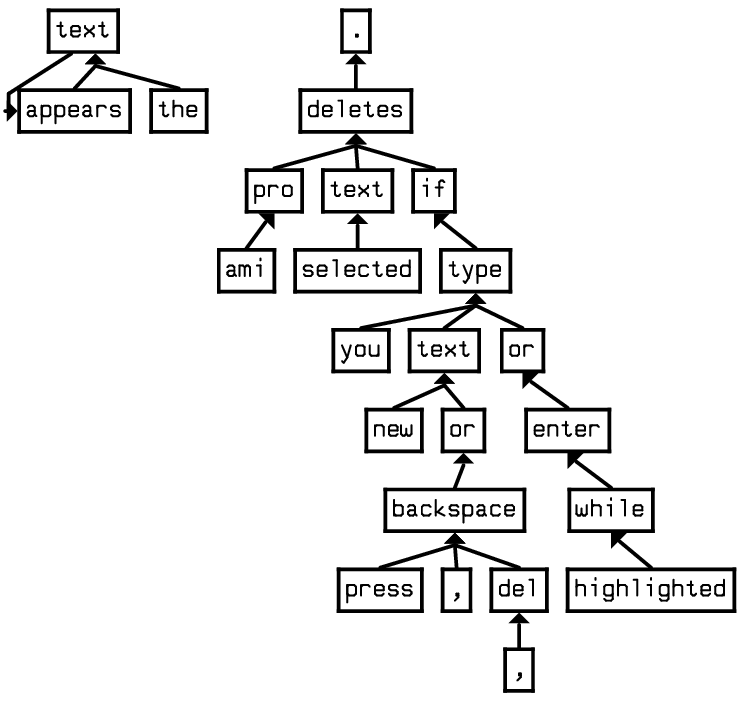}}}
\hfill
\end{center}
\begin{center}
\parbox{\textwidth}{
{\bf Figure~7. The parse tree produced by the parser enhanced
with the divide-and-conquer strategy on this sentence in the Lotus data
set -- {\sf ``Ami Pro deletes selected text if you type new text or press
BACKSPACE, DEL, or ENTER while the text appears highlighted."}.}
The disambiguation phase wrongly classifies the {\sf ``or"} before {\sf ``press"}
as a logical conjunction. It also wrongly classifies the {\sf ``or"}  before
{\sf ``ENTER"} as a clausal conjunction. The reason behind this error in 
disambiguation is the erroneous parts of speech assigned to 
{\sf ``press"} and {\sf ``ENTER"} by the tagger -- {\sf ``press"} is tagged as
a proper noun whilst {\sf ``ENTER"} is tagged as a verb. These disambiguation
errors result in an error in the segmentation of the sentence into 
{\sf ``Ami Pro deletes selected text."}, {\sf ``You type new text or press
BACKSPACE, DEL."}, {\sf ``ENTER while the text appears highlighted."} and
propagated down to the synthesis phase too. The cyclic part of the parse 
tree which was present in the original parser's output was not eradicated 
here.
}
\end{center}
\end{figure}

\section{Concluding Remarks}
\label{sec-conclusion}

In this paper, we have shown that the accuracy of the parser
can be considerably improved when presented with simple sub-sentences instead
of long, complex sentences. The methodology proposed does not require 
enhancements to the parsing mechanism itself, and hence is generic enough
to be applicable to any dependency or constituency structure parsers.

The divide-and-conquer strategy may be further improved. The disambiguation
phase may rely less on the performance of the tagger if more discriminating
features, in addition to the parts of speech, are used. Other punctuations
such as colons and semi-colons can also be incorporated. As for the
synthesis engine, the rule-based approach may be replaced with a more
portable statistical approach.

It should also be noted that the divide-and-conquer strategy need not be 
constrained to the enhancement of parsing. 
Its disambiguation and segmentation phases may actually
be adapted to other areas of NLP, such as generation.


\begin{thebibliography}{}

\bibitem[\protect\citename{Black}1993]{Black:state-of-the-art}
Ezra Black.
\newblock 1993.
\newblock Parsing English By Computer: The State Of The Art.
\newblock {\em Internal report}, ATR Interpreting Telecommunications Research Laboratories. 

\bibitem[\protect\citename{Briscoe and Carroll}1995]{Briscoe+Carroll:punc}
Ted Briscoe and John Carroll.
\newblock 1995.
\newblock Developing and Evaluating A Probabilistic LR Parser of Part-of-Sppech and Punctuation Labels.
\newblock In {\em Proceedings of the ACL/SIGPARSE 4th International Workshop on Parsing Technologies}, Prague / Karlovy Vary, Czech Republic, pages 48--58.

\bibitem[\protect\citename{Jones}1994]{Jones:punc}
Bernard E. M. Jones.
\newblock 1994.
\newblock Exploring the Role of Punctuation in Parsing Natural Text.
\newblock In {\em Proceedings of COLING-94,} pages 421--425.

\bibitem[\protect\citename{Kurohashi and Nagao}1994]{Kurohashi+Nagao:conj}
Sadao Kurohashi and Makoto Nagao.
\newblock 1994.
\newblock A Syntactic Analysis Method of Long Japanese Sentences Based on the
Detection of Conjunctive Structures.
\newblock {\em Computational Linguistics}, 20(4):507--534.

\bibitem[\protect\citename{Magerman}1994]{Magerman:spatter}
David M. Magerman.
\newblock 1994.
\newblock Natural Language Parsing as Statistical Pattern Recognition.
\newblock {\em PhD thesis,} Stanford University.

\bibitem[\protect\citename{Mel'\v{c}uk}1988]{Melcuk:dep}
Igor A. Mel'\v{c}uk. 
\newblock 1988.
\newblock Dependency Syntax : Theory and Practice.
\newblock {\em State University of New York Press}. 

\bibitem[\protect\citename{Marcus \bgroup et al.\egroup}1993]{Marcus+al:penntree}
Mitchell P. Marcus, Beatrice Santorini, Mary A. Marcinkiewicz. 
\newblock 1993.
\newblock Building a Large Annotated Corpus of English: The Penn Treebank.
\newblock {\em Computational Linguistics}, 19(2):313--330.

\bibitem[\protect\citename{Nunberg}1990]{Nunberg:punc}
Geoffrey Nunberg.
\newblock 1990.
\newblock The Linguistics of Punctuation.
\newblock {\em CSLI Lecture Notes}, 18. 

\bibitem[\protect\citename{Ting}1995]{Ting:npparser}
Christopher H. A. Ting.
\newblock 1995.
\newblock Parsing of Noun Phrases with HMM.
\newblock In {\em Proceedings of Inter-Faculty Seminar on Natural Language 
Processing}, National University of Singapore.

\bibitem[\protect\citename{Ting and How}1995]{Ting+How:ipsm}
Christopher H. A. Ting and K. Y. How.
\newblock 1995.
\newblock Parsing Dependency Structures Without Using Any Grammar Formalism.
\newblock In {\em Proceedings of the International Parser Systems Workshop' '95}, University of Limerick, Ireland.

\bibitem[\protect\citename{Ting and Peh}1995]{Ting+Peh:ipsm}
Christopher H. A. Ting and L. S. Peh.
\newblock 1995.
\newblock DESPAR, A Dependency Structure Parser Without Using Any Grammar
Formalism. 
\newblock In Richard F. E. Sutcliffe and Heinz-Detlev Koch, editors, {\em Industrial Parsing of Software Manuals}. Editions Rodopi B. V., Amsterdam.
\newblock To appear.

\end{thebibliography}
\end{document}